\documentclass[runningheads]{llncs}

\usepackage{eccv}

\usepackage{eccvabbrv}

\usepackage{graphicx}
\usepackage{lipsum}
\usepackage{multirow}
\usepackage{booktabs}
\usepackage{pifont}
\usepackage{wrapfig}
\usepackage{amsmath}
\usepackage{tabularray}
\usepackage{adjustbox}

\usepackage[accsupp]{axessibility}  

\usepackage{hyperref}

\usepackage{orcidlink}

\begin{document}



\title{SCP-Diff: \underline{S}patial-\underline{C}ategorical Joint \underline{P}rior for \underline{Diff}usion Based Semantic Image Synthesis} 

\titlerunning{SCP-Diff}

\author{
Huan-ang Gao\textsuperscript{*}\inst{1} \and
Mingju Gao\textsuperscript{*}\inst{1} \and
Jiaju Li\inst{1, 2} \and
Wenyi Li\inst{1} \and \\
Rong Zhi\inst{3} \and
Hao Tang\inst{4} \and
Hao Zhao\textsuperscript{\textdagger}\inst{1}
}

\authorrunning{Gao et al.}

\institute{Institute for AI Industry Research (AIR), Tsinghua University \and University of Chinese Academy of Sciences \and
Mercedes-Benz Group China Ltd.  \and
Peking University \& Carnegie Mellon University \\
\email{gha24@mails.tsinghua.edu.cn},\ \email{mgao18926@gmail.com} \\ \email{zhaohao@air.tsinghua.edu.cn}
}

\maketitle
\setcounter{footnote}{1}
\footnotetext{\textsuperscript{*} Indicates Equal Contribution. \textsuperscript{\textdagger} Indicates Corresponding Author.}
\setcounter{footnote}{2}
\footnotetext{Project Page: \url{https://air-discover.github.io/SCP-Diff/}}

\begin{figure}[h]
\centering
\includegraphics[width=0.75\textwidth]{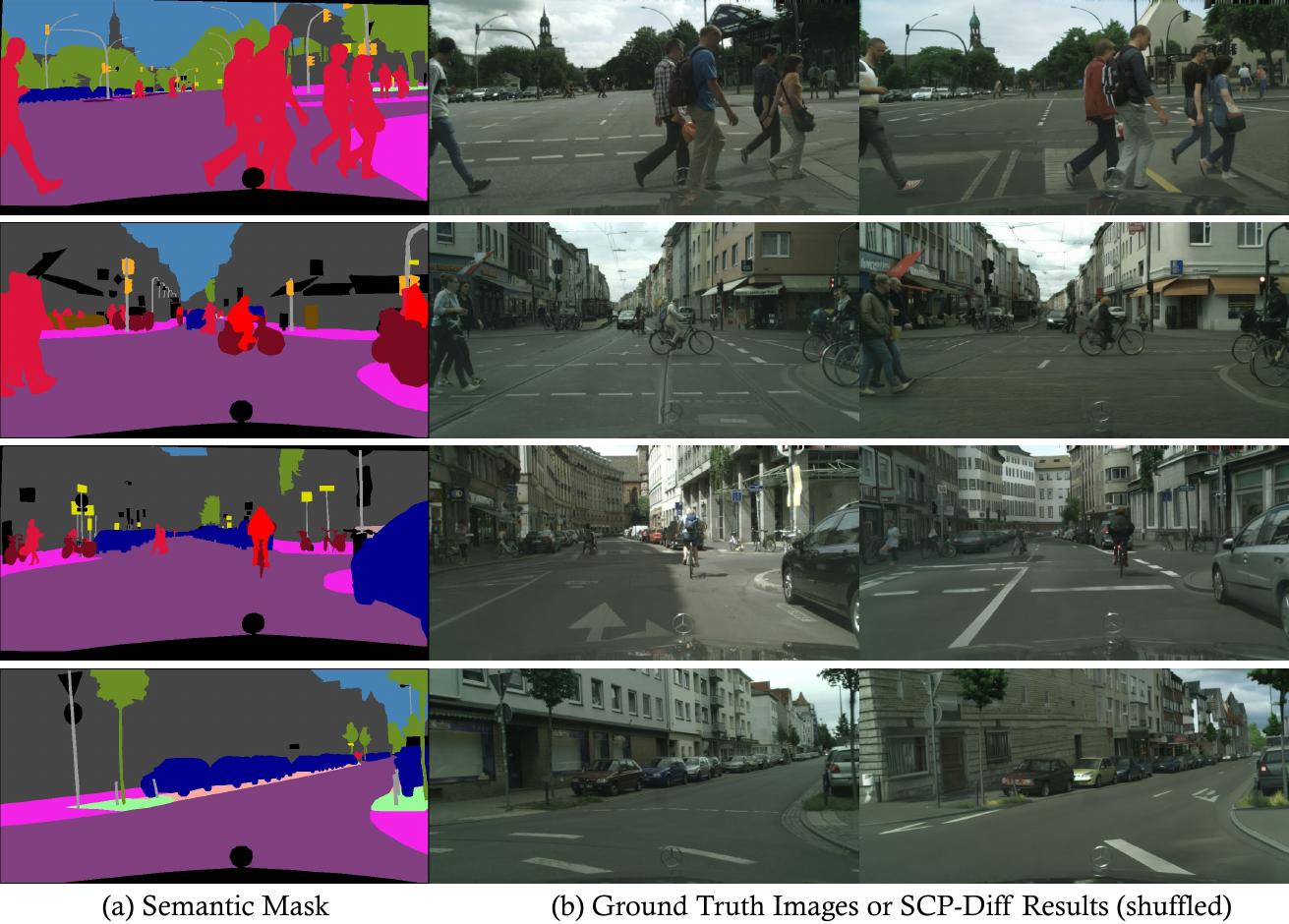}
\caption{\textbf{For single-domain Cityscapes \cite{cordts2016cityscapes}, our method can generate photo-realistic images from semantic masks (a). In (b), we shuffle ground truth real images and SCP-Diff results and the right answer is in the footnote of the conclusion. While the state-of-the-art method ECGAN \cite{tang2023edge} achieves 44.5 FID on Cityscapes, our method achieves 10.5 FID. The quality is credited to the strong spatial and categorical prior of Cityscapes.}}
\label{fig:cs-teaser}
\end{figure}

\begin{abstract}

Semantic image synthesis (SIS) shows good promises for sensor simulation.
However, current best practices in this field, based on GANs, have not yet reached the desired level of quality.
As latent diffusion models make significant strides in image generation, we are prompted to evaluate ControlNet, a notable method for its dense control capabilities.
Our investigation uncovered two primary issues with its results: the presence of weird sub-structures within large semantic areas and the misalignment of content with the semantic mask.
Through empirical study, we pinpointed the cause of these problems as a mismatch between the noised training data distribution and the standard normal prior applied at the inference stage.
To address this challenge, we developed specific noise priors for SIS, encompassing spatial, categorical, and an innovative \underline{\textbf{s}}patial-\underline{\textbf{c}}ategorical joint \underline{\textbf{p}}rior for inference. 
This approach, which we have named \textbf{SCP}-Diff, has set new state-of-the-art results in SIS on Cityscapes, ADE20K and COCO-Stuff, yielding a FID as low as 10.53 on Cityscapes.
The code and models can be accessed via the project page.

\end{abstract}

\section{Introduction}
\label{sec:intro}

\begin{figure}[t]
\centering  
\includegraphics[width=0.75\textwidth]{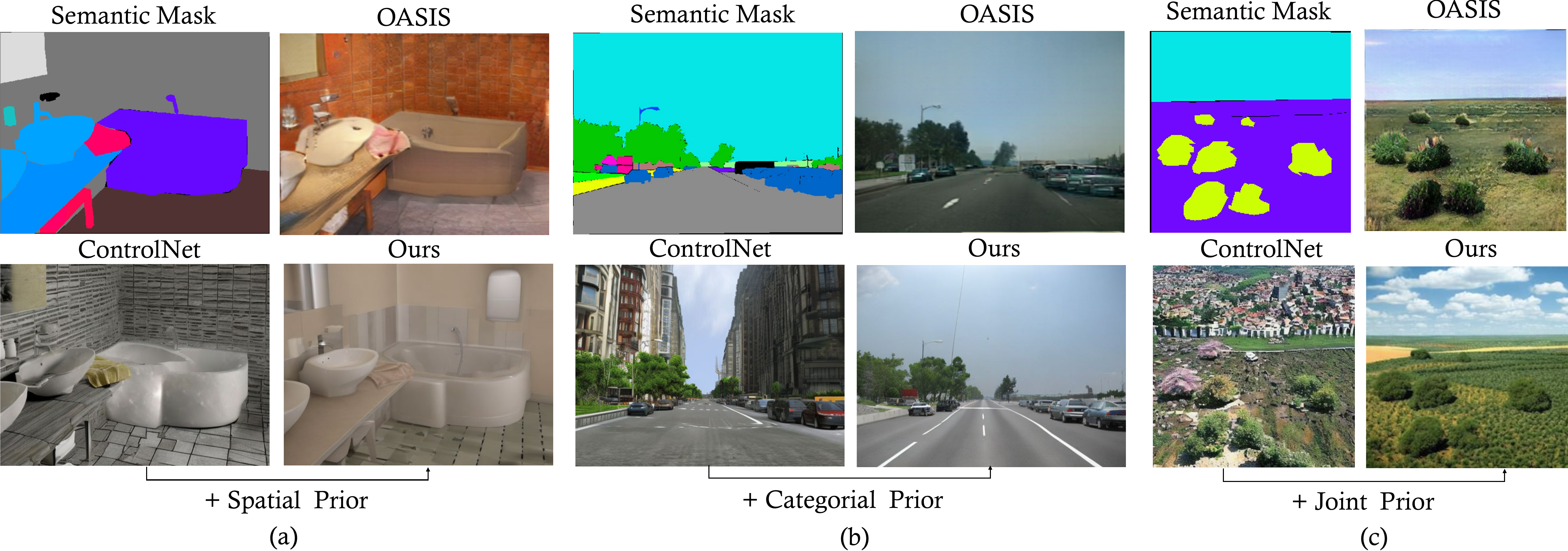}
\caption{
\textbf{We explain the underlying mechanisms of our proposed noise priors using ADE20K} \cite{zhou2017scene}\textbf{.}
(a) Introducing the spatial prior aligns the image style with the dataset's aesthetic while removing odd substructures in large semantic areas. (b) Incorporating the class prior can enhance alignment with the provided semantic masks. (c) By jointing spatial and class priors, their beneficial features are combined, allowing our joint prior (SCP-Diff) to achieve state-of-the-art results on ADE20K.
}
\label{fig:ade-teaser}
\end{figure}

Semantic image synthesis is dedicated to generating high-quality images aligned with semantic maps, offering users the ability to accurately control the spatial layout of the generated images. 
The potential of this technology is particularly notable in fields like autonomous driving and robotics, because it enables the creation of highly realistic and diverse virtual environments with semantic layout control, reducing the dependency on extensive real-world data collection \cite{luo2022context, luo2023siedob, ntavelis2020sesame, yang2024freemask, zheng2023steps} and boosting the performance for perception modules \cite{gao2023semi, gao2023dqs3d, tian2023unsupervised, jiang2024p}.

However, current leading techniques \cite{sushko2022oasis, tang2023edge} in this area, which rely on Generative Adversarial Networks (GANs), have not reached the anticipated quality levels necessary for the practical application of sensor simulation. 
As latent diffusion models \cite{rombach2022high,chen2024ultraman,li2024fairdiff} achieve remarkable success in image creation, we are motivated to evaluate ControlNet \cite{zhang2023adding}, a method that enables dense control over Stable Diffusion \cite{rombach2022high}.
We finetuned ControlNet with semantic label masks as conditions, but identified two significant issues in the generated results: (i) the emergence of weird sub-structures in large semantic regions (Fig.~\ref{fig:ade-teaser}(a), the bathtub is divided into two parts) and (ii) the misalignment of content placement with the provided semantic masks (Fig.~\ref{fig:ade-teaser}(b), buildings appear in the sky region).

Why do large-scale pretrained latent diffusion models struggle to complete this semantic image synthesis task after finetuning? Our empirical analysis (see Fig.~\ref{fig:x0_prior}) reveals that the primary source of discrepancy in quality between the generated outcomes and the actual images is not from the score matching learning (\ie, finetuning) process. Instead, it originates from a mismatch \cite{lin2024common} between the distribution of noised data used during training and the standard normal distribution typically employed during the inference process.

\textbf{Devise noise priors for inference.}
To address the mismatch in the distributions between training and inference, we implemented inference-time noise priors tailored for SIS, which can be seamlessly integrated into the finetuned ControlNet without further training.
Initially, we introduced a spatial prior by estimating the distribution of real latents through a Gaussian model and averaging the outcomes across batches, similar to \cite{everaert2024exploiting}. This approach notably improved the organization of scene layouts (for instance, the bathtub is now recognized as a single entity rather than fragmented parts in Fig.~\ref{fig:ade-teaser}(a)) and enriched the diversity of colors observed in the images. 
This prior is closer to the training trajectory; however, discrepancies with the provided label masks remained evident (such as an unintended lamp appearing on the wall in Fig.~\ref{fig:ade-teaser}(a)).
We argue that this issue arises because the spatial noise prior unfits the control branch of ControlNet, where the former contains a mixture of modes (corresponding to different categories) during reduction, which may hinder the latter to generate meaningful residuals (added back to the SD branch) for \textit{on-trajectory} denoising.
To further refine our approach, we explored a categorical prior by aggregating real-image latents by class and started denoising from the aggregated statistics. This strategy helped reduce label alignment issues, though it resulted in the outputs reverting to a monotonous color scheme, as shown in Fig.~\ref{fig:ade-teaser}(b).

To combine the best of both worlds, we introduced the \underline{\textbf{S}}patial-\underline{\textbf{C}}ategorical Joint \underline{\textbf{P}}rior and coined this \underline{\textbf{diff}}usion-based synthesis technique \textbf{SCP-Diff}.
Our thorough experiments have confirmed the efficacy of SCP-Diff. Remarkably, despite its simplicity and the fact that it is training-free, SCP-Diff sets state-of-the-art on single-domain Cityscapes and multi-domain ADE20K simultanously when applied to ControlNet, synthesizes photo-realistic scenes that are difficult to distinguish from real ones (see Fig.~\ref{fig:cs-teaser}, \ref{fig:ade-teaser}(c)).
It also sets state-of-the-art synthesis result on COCO-Stuff when applied to FreestyleNet \cite{xue2023freestyle}, previous state-of-the-art diffusion-based SIS approach on COCO-Stuff.
Aside from quality, we also quantitatively demonstrate that incorporating the prior does not negatively impact the diversity of diffusion-based generative modeling.
User studies are also supportive for the quality and fidelity of the generated images to the provided label masks.
The contributions of this work can be summarized as follows:
\begin{itemize}
    \item We pinpoint the challenge posed by the discrepancy in inference distribution in finetuned ControlNets for Semantic Image Synthesis (SIS) and introduce a solution that utilizes inference noise priors to bridge this gap, notably without the need for retraining.
    \item We elucidate the design philosophy behind the creation of inference noise priors tailored for the SIS task, unpacking the mechanics of spatial and categorical priors within SIS, and finally integrate them into a joint one.
    \item Demonstrating superior capabilities, our integrated joint prior (SCP-Diff) achieves state-of-the-art performance on three well-established SIS benchmarks, Cityscapes, ADE20K and COCO-Stuff, delivering high-quality outputs that are hard to distinguish from the real-world images.
\end{itemize}



\section{Related Work}

\subsection{Semantic Image Synthesis}
Semantic image synthesis (SIS) \cite{lu2019closed, park2019semantic, wang2018high, tang2019multi, tang2020local, lv2024place} aims to create photo-realistic images from semantic label maps. 
Previous works in this field are mostly based on Generative Adversarial Networks (GANs), which are trained using both adversarial loss \cite{goodfellow2020generative} and reconstruction loss. 
A key advancement in network design was introduced with AdaIN \cite{huang2017arbitrary}, which aligns the mean and variance of content features with those of the style features.
Building on this modulation concept, SPADE \cite{park2019semantic} proposes spatially adaptive normalization to better embed semantic layouts into the generator.
CLADE \cite{tan2021efficient} further optimized SPADE's approach by introducing a more efficient normalization layer that adapts to different semantic classes.
The state-of-the-art GAN-based approach, ECGAN \cite{tang2023edge}, proposes to use the edge as an intermediate representation to help modulate the image generation process and employs contrastive learning to derive embeddings rich in semantic information.

The advent of conditional denoising diffusion probabilistic models \cite{ho2020denoising, song2020denoising, duan2023diffusiondepth} has spurred innovations in SIS.
FreestyleNet \cite{xue2023freestyle} innovates within this space by adjusting the cross-attention maps, ensuring that each text token influences only the pixel regions delineated by the semantic mask.
Following the modulation concept, SDM \cite{wang2022semantic} designs a novel denoiser architecture with modulation layers.
However, these methods directly operate in pixel space and, consequently, produce outcomes that are not as high-resolution as those generated by GAN-based techniques due to VRAM constraints.

\subsection{Latent Diffusion-based Controllable Generation}
Addressing the challenges of slow inference speeds and exceedingly high training expenses in pixel space, Stable Diffusion \cite{rombach2022high} introduced a two-stage image synthesis strategy. Initially, it employs VQGANs \cite{esser2021taming} to compress the image into a discretized latent code. Subsequently, it utilizes powerful diffusion-based generative modeling to process these latent representations, employing a UNet denoiser to exploit the inductive biases in images, which supports the text condition via cross attention.
Building upon Stable Diffusion \cite{rombach2022high}, numerous efforts aim to broaden control to additional modalities, such as incorporating CLIP features \cite{ramesh2022hierarchical}, or modifying specific regions through the inpainting formulation \cite{avrahami2022blended, couairon2022diffedit}.

ControlNet \cite{zhang2023adding} offers task-specific control by using a control branch to encode conditions.
Within SIS, these conditions can be specified by color-coded semantic masks, and we report that such formulation can achieve state-of-the-art performance over previous SIS methods.
Nevertheless, a more thorough examination uncovers issues such as the generation of unrealistic sub-structure and misalignment between the generated images and the corresponding semantic masks.
This paper posits that the root cause of these issues lies in the initial noise setup and suggests the integration of the pre-computed noise priors. 

\subsection{Playing Noise Tricks in Diffusion Models}

\textbf{Sampling Inversion.} As DDIM sampling \cite{song2020denoising} operates deterministically, Null-text Inversion \cite{mokady2023null} learns to inverse this sampling process to extract noise from earlier timesteps, which is ideal for editing the signal image but not for generating new ones with label masks.

\textbf{Signal Leakage in the Noised Latents.}
In video diffusion models, PYoCo \cite{ge2023preserve} observed that the noise maps corresponding to frames from the same video tend to group together.
FreeNoise \cite{qiu2023freenoise} shows that rescheduling a sequence of noises enables long-range correlation modeling for longer video generation.
These findings point to the assumption that common diffusion noise schedules inadequately corrupt signals in images, as discussed in \cite{everaert2024exploiting}.
To address this, FreeInit \cite{wu2023freeinit} refines the low-frequency components of the inference initial noise in an iterative manner in video diffusion models at the cost of inference time. \cite{lin2024common} recommends retraining with a different noise schedule, which is computationally intensive.
Note that none of these works specifically addresses SIS, where the key challenge is maintaining alignment with labels for controlled generation.

\section{Preliminaries and Observations}

\subsection{Preliminaries}

Given a semantic segmentation mask $M \in \mathbb{N}^{H \times W}$, where $H$ and $W$ represent height and width, and each element within $M$ corresponds to a semantic label assigned to a specific pixel, the objective of semantic image synthesis (SIS) is to devise a function that transforms $M$ into a photorealistic image $I \in \mathbb{R}^{H \times W \times 3}$.

The state-of-the-art method for SIS, ControlNet \cite{zhang2023adding}, is trained to reconstruct an initial image $x_0$ by removing noise from a distorted image $x_t$ in a given timestep $t \in [0, T]$. 
At each timestep $t$, the denoising model $\boldsymbol{\varepsilon}_\theta$ is presented with,
\begin{equation}
    \label{eq:adding_noise}
    x_t = \sqrt{\alpha_t}x_0 + \sqrt{(1 - \alpha_t)}\varepsilon_t,
\end{equation}
where $x_0$ is the image encoded by VQGAN \cite{esser2021taming} encoder, $\varepsilon_t \sim \mathcal{N}(0, I)$ and $\alpha_t$ is the cumulative product of scaling at each timestep $t$. 
Taking $x_t$ as input, the denoising model predicts the added noise $\hat{\varepsilon}_t = \boldsymbol{\varepsilon}_\theta(x_t, t, T, M)$, where $T$ is the encoding of the text condition and $M$ is the label mask.
Within the model, a non-trainable branch retains its configuration from the pretrained weights of Stable Diffusion \cite{rombach2022high} and is tasked with processing $x_t, t,$ and $T$. Simultaneously, the model duplicates its encoder to a separate, trainable branch to process $M$, which is subsequently linked to the non-trainable one using layers of zero convolution.
The loss function used to train the denoiser is defined as,
\begin{equation}
    \label{eq:training}
    \mathcal{L} = \mathbb{E}_{x_0, t, T, M, \varepsilon_t}[\|\varepsilon_t - \hat{\varepsilon}_t\|^2].
\end{equation}

At inference time, we sample $x_T \sim \mathcal{N}_\text{normal}$, \ie, $\mathcal{N}(0, I)$ and conduct the reverse denosing process iteratively from $t=T$ until $t=1$, 
\begin{equation}
    x_{t-1} = \sqrt{\alpha_{t-1}} \left( \frac{x_t - \sqrt{1 - \alpha_t} \cdot \hat{\varepsilon}_t}{\sqrt{\alpha_t}} \right) + \sqrt{1 - \alpha_{t-1}} \cdot \hat{\varepsilon}_t.
\end{equation}
After $T$ iterations, we get $x_0$ and decode it to $I$ using the VQGAN decoder \cite{esser2021taming}.

\subsection{Denoising from $\mathcal{N}_\text{normal}$? An Unreliable Inference Assumption}

\begin{wrapfigure}[17]{r}{0.45\textwidth}
	\begin{minipage}[t]{0.45\textwidth}
    \includegraphics[width=\linewidth]{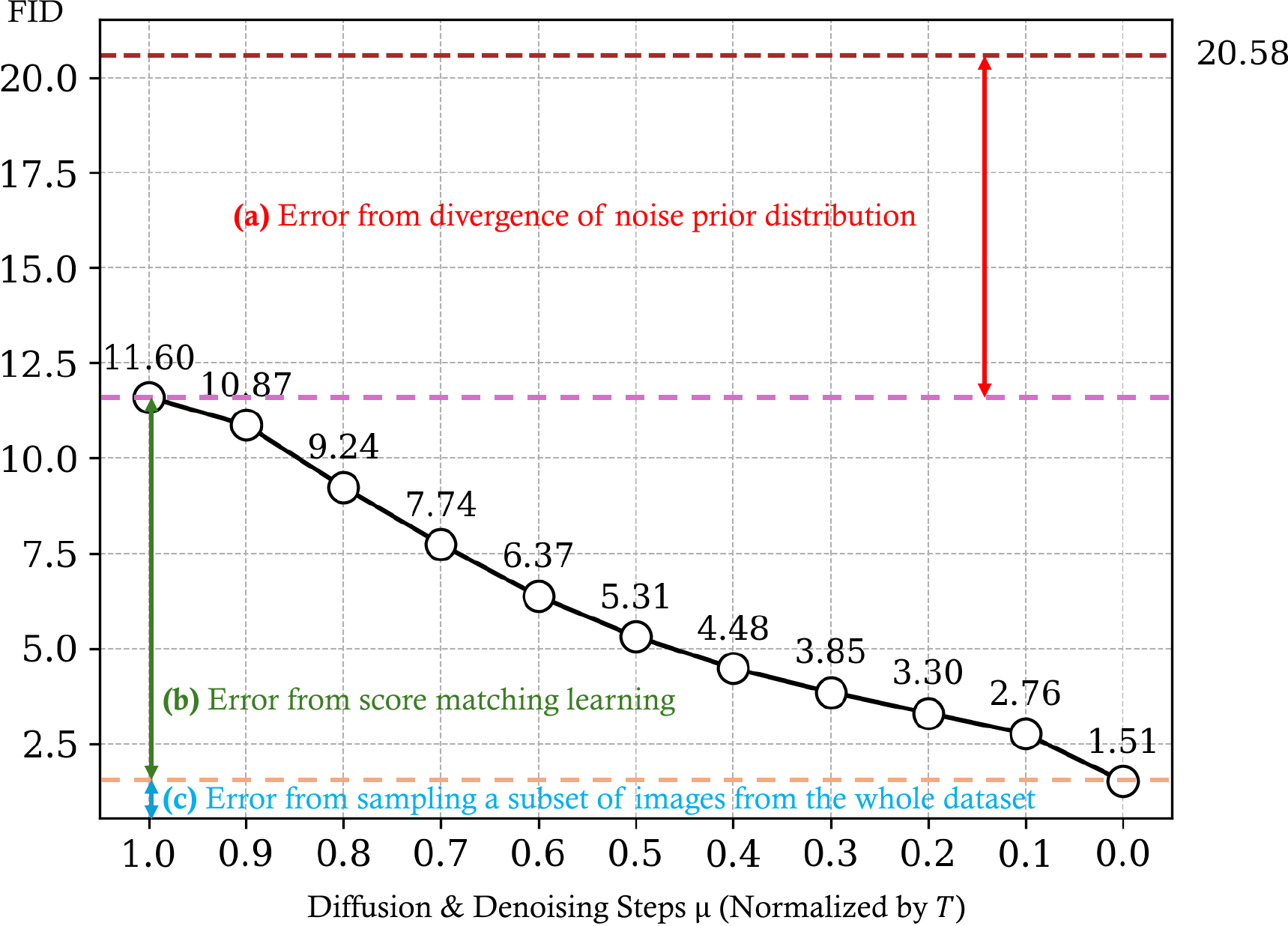}
    \caption{An empirical study of denoising priors. \textcolor{brown}{Brown dotted} line denotes the FID denoising from $\mathcal{N}_\text{normal}$, and the polyline illustrates the FID denoising from $\mathcal{N}_{x_0, \mu T}$ introduced in main text.}
    \label{fig:x0_prior}
	\end{minipage}
\end{wrapfigure}

Simply applying ControlNet \cite{zhang2023adding} finetuning to the Semantic Image Synthesis (SIS) task results in suboptimal outcomes, 
such as monotonous color schemes, weird substuctures, or an inability to accurately follow semantic label masks, 
as illustrated in Figs.~\ref{fig:ade-teaser} and \ref{fig:prior-comp}. 
To understand the underlying reasons for these issues, we perform an empirical analysis by finetuning on the ADE20K dataset.
The results are reported in Fig.~\ref{fig:x0_prior}, where we try to analyze the errors associated with the generated results compared to the images from the ADE20K dataset.

We introduce $\mathcal{N}_{x_0, \mu T}$ as a means to investigate whether the score matching learning process, specifically the optimization of Eq.~\eqref{eq:training}, incorporates any errors,
\begin{equation}
    \mathcal{N}_{x_0, \mu T} := \mathcal{N}(\sqrt{\alpha_{\mu T}}x_0, (1-\alpha_{\mu T}) I),
\end{equation}
Intuitively, for each pair of $(x_0, M)$, we supply the denoising model with $x_{\mu T}$ (as defined in Eq.~\eqref{eq:adding_noise}, $\mu \in [0, 1]$) and the condition branch with $M$, followed by assessing the resultant generation using the FID score. 
Although an increase in the FID score is expected with the progression of the denoising steps ${\mu T}$ (Fig.~\ref{fig:x0_prior}(b)), a significant discrepancy (Fig.~\ref{fig:x0_prior}(a)) remains between the denoising from $\mathcal{N}_{x_0, T}$ ($\mu = 1$) and a standard denoising inference process, $\mathcal{N}_\text{normal}$.

The empirical study reveals that the common inference assumption that $x_T$ should closely resemble $\mathcal{N}_\text{normal}$ is unreliable. 
This argument is also supported by the analysis presented in \cite{lin2024common, zhang2024preserving}, arguing that current settings of $\alpha_t$ in practice can make the denoising model distinguish between initialization samples (namely $x_T$ in our study) in training and testing cases.
Nevertheless, their proposed solution, which involves retraining diffusion models, is costly and does not leverage important domain knowledge of semantic image synthesis, which demands (i) a comprehensive grasp of scene layouts from segmentation masks and (ii) adherence to pixel-wise dense semantic label maps at the same time.

\begin{figure}[t]
\centering
\includegraphics[width=0.8\linewidth]{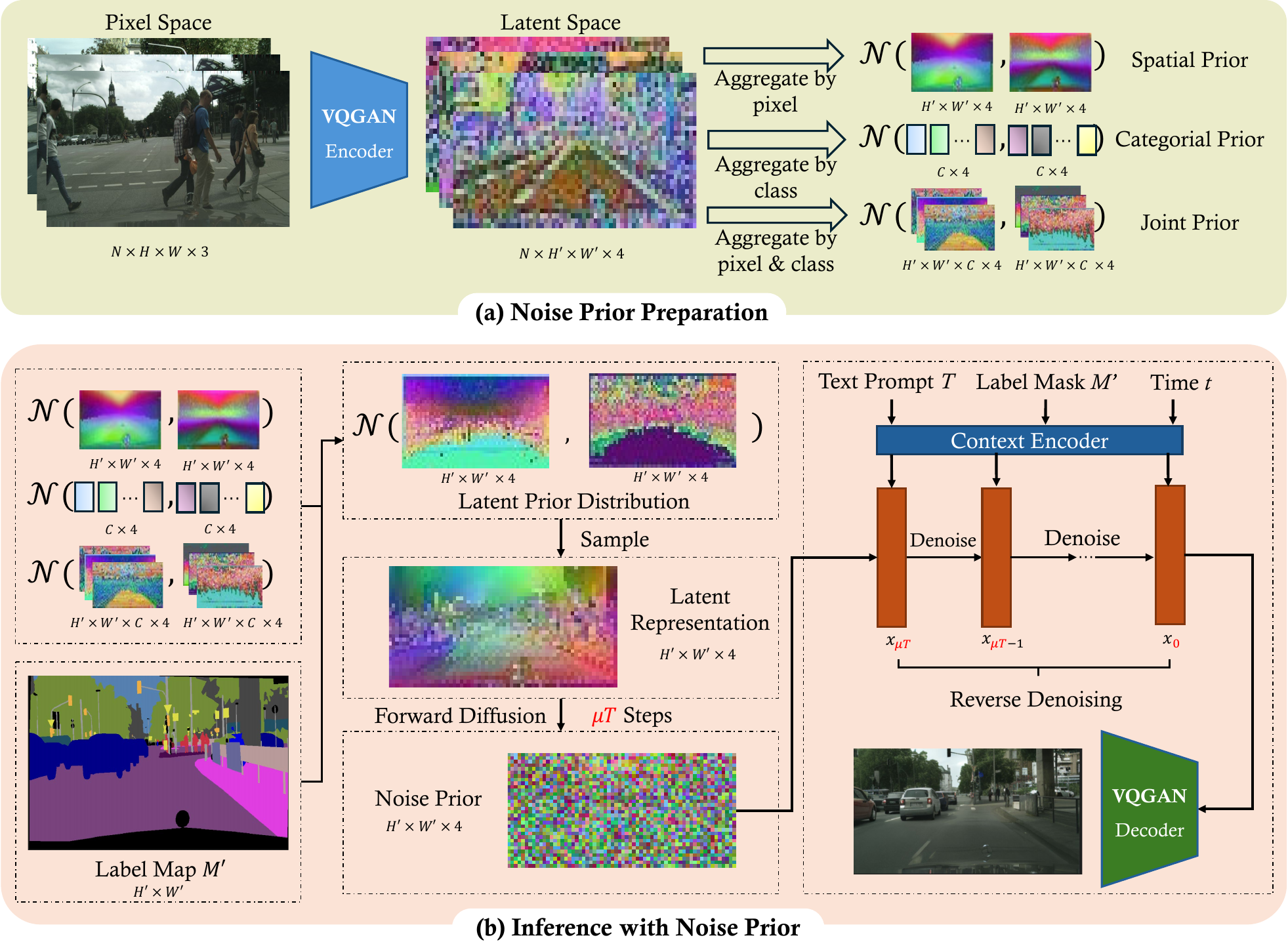}
\caption{\textbf{Overview of our proposed framework. $\mathcal{N}$ stands for Gaussian distribution.}}
\label{fig:main}
\end{figure}

\section{Method}
\subsection{Overview}

In Fig.~\ref{fig:main}, we provide an overview of our proposed framework, which is structured into two main stages: the preparation of latent priors and the application of these pre-computed latent priors during inference.

\textbf{Noise Prior Preparation.} With a set of $N$ reference images, this phase involves reducing these images into latent priors, approximated by Gaussian distributions. Initially, the images are transformed into the latent space using a pretrained VQGAN \cite{esser2021taming} encoder, followed by computing the means and variances. Depending on the type of prior needed, we apply Eq.~\eqref{eq:spatial_prior} for the spatial prior, Eq.~\eqref{eq:class_prior} for the categorical prior, and Eq.~\eqref{eq:joint_prior} for the joint prior.

\textbf{Inference with Noise Prior.} We first assemble the latent prior distribution map aligned with the provided downsampled label map $M'$.
For the spatial prior, we directly replicate it. Otherwise, $M'$ is used to index the specified prior on a token-by-token basis to construct a distribution map. 
From this distribution, we sample a latent representation and introduce noise for $\mu T$ steps to create the noise prior. 
This noise prior is then processed through the fine-tuned ControlNet \cite{zhang2023adding}, which denoises the last $\mu T$ steps. The final step involves utilizing the pretrained VQGAN \cite{esser2021taming} decoder to reconstruct the generated image.

\subsection{Exploring Spatial Prior and Categorical Prior}
\label{spatial-class-prior}
With $N$ reference latent images and their associated masks $\{(x^{(i)}_0, M^{(i)})\}_{i=1}^N$ from the dataset, our objective is to reduce them into noise priors $\mathcal{N}_{x_{\mu T}}$ which aligns with the training trajectory of noise distributions, facilitating low-error inference.

\begin{figure}[t]
\centering
\includegraphics[width=0.7\textwidth]{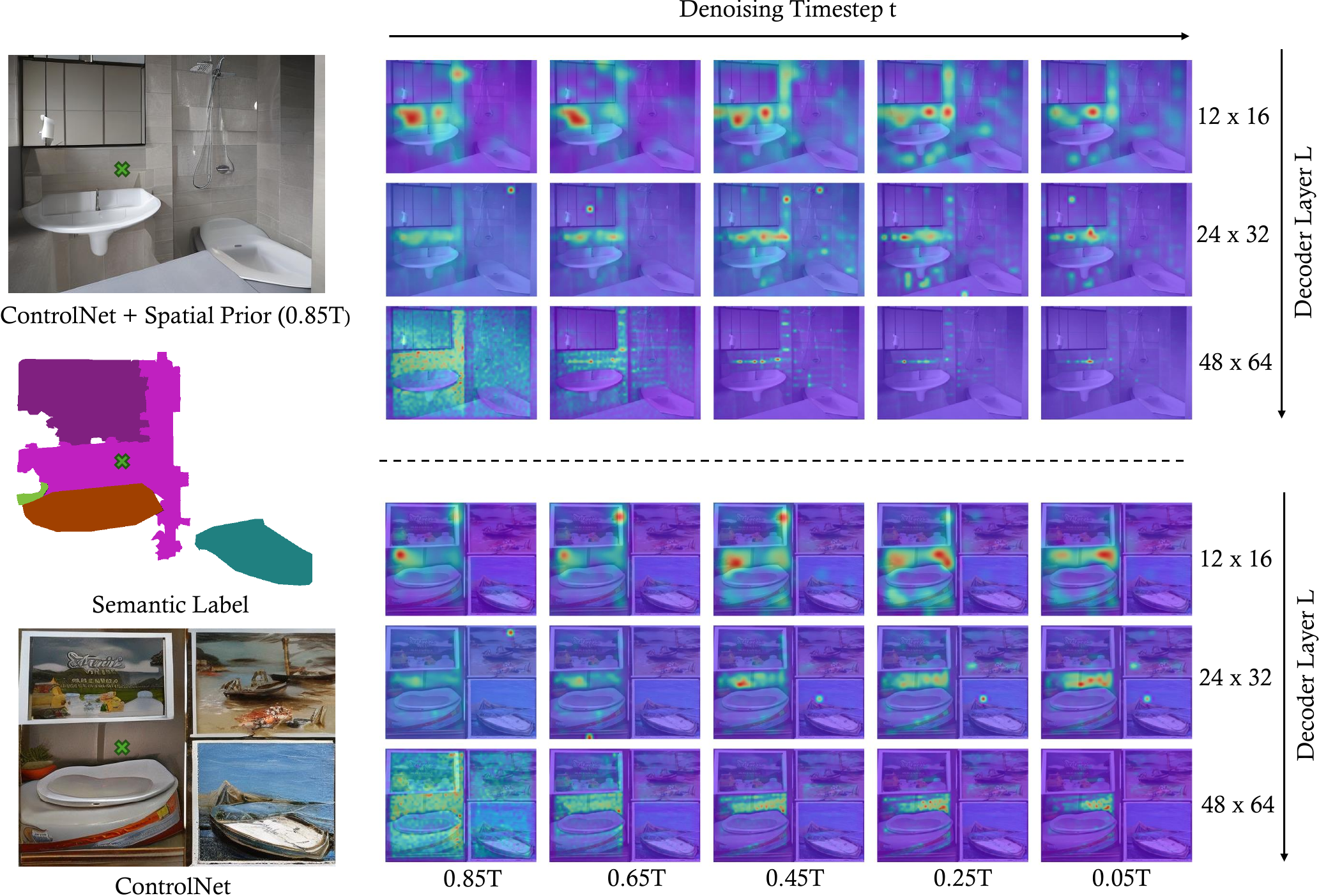}
\caption{\textbf{Case study on spatial prior.} We provided the finetuned ControlNet \cite{zhang2023adding} with identical semantic label maps (middle on the left, where white pixels represent 'unlabeled'), but introduced different priors, spatial prior (above) and normal prior (below). We visualize attention maps of decoder blocks where the query region is highlighted with a `\textcolor{green}{\ding{53}}'. The images beneath the attention map masks represent the ultimate decoding outcomes and are included solely for visualization purposes.
}
\label{fig:spatial-prior}
\end{figure}

\textbf{Spatial Prior.}
Given that the training process is focused on minimizing the score matching loss (Eq.~\eqref{eq:training}) for each instance of $x_0$ in expectation, with the theoretical premise that every $x_0$ has an equal impact on the ultimately learned training noise distribution, our initial step is to diminish the reference latent images $\{x^{(i)}_0\}_{i=1}^N$ across the batch dimension.

More specifically, we define the spatial prior as,
\begin{equation}
\label{eq:spatial_prior}
    \mathcal{N}_\text{spatial} := \mathcal{N}\left(\sum_{i=1}^N \frac{x_0^{(i)}}{N}, \left[\sum_{i=1}^N \frac{1}{N}  {\left[ 
x_0^{(i)} - \mu(x_0^{(i)}) \right] ^2}\right] \odot I_{H' \times W' \times 4}\right),
\end{equation}
where $\odot$ denotes the Hadamard product. For the sake of simplification in our reduction process, we choose not to model the correlations between spatial tokens collectively, treating each spatial location as an individual marginal distribution instead. For initializing noise values for the inference process started at $\mu T$, we sample from,
\begin{equation}
    \label{eq:spatial-uT}
    \mathcal{N}_{\text{spatial}, \mu T} := \mathcal{N}\left(\sqrt{\alpha_{\mu T}} \cdot x_\text{spatial}, (1-\alpha_{\mu T}) I \right), \ \text{where} \  x_\text{spatial} \sim \mathcal{N}_\text{spatial}.
\end{equation}

So, what is encoded by the spatial prior? To answer this, we conducted a case study, illustrated in Fig.~\ref{fig:spatial-prior}. From the analysis, we see that the group using spatial priors exhibits a broader receptive field in constructing scene layouts, in contrast to the group employing normal priors, which quickly focuses its attention narrowly on local fields. This distinction sheds light on why the spatial prior group can generate completed scenes with less weird sub-structures, while the normal prior group's output resembles cropping and pasting objects following similar shape masks.
Further insights from Fig.~\ref{fig:prior-comp} demonstrate that the use of spatial priors facilitates the production of images that are consistent with the dataset style and enriched with a wider spectrum of colors and textures.

\begin{figure}[t]
\centering
\includegraphics[width=0.75\textwidth]{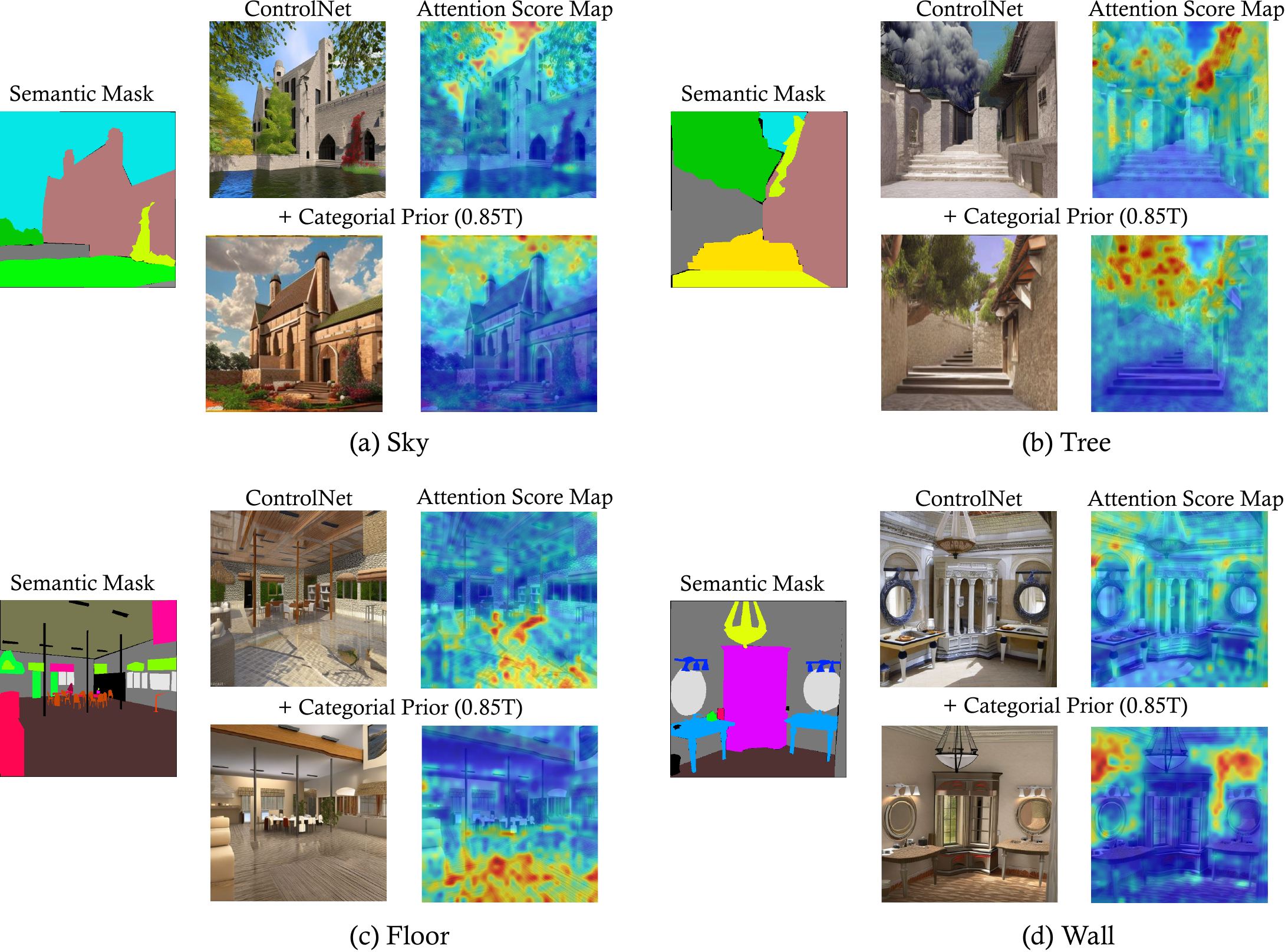}
\caption{\textbf{Case study on categorial prior. }
We examine the intermediate features of encoder blocks from the middle of the denoising process where the object shapes begin to form \cite{voynov2023p+, balaji2022ediffi, cao2023masactrl}. 
The cross-attention score map is computed for these features against the language embeddings corresponding to various categories. It's important to note that this extra categorical information from natural language is utilized solely for visualization and does \textit{not} used in the actual denoising process. The denoiser exclusively identifies categories through the semantic mask and our categorical prior.
}
\label{fig:categorical-prior}
\end{figure}

\textbf{Categorical Prior.}
While spatial priors succeed in achieving global attention across the scene to construct scene layouts with fidelity, they fall short in incorporating class-specific information. 
This shortfall becomes apparent through hallucinatory artifacts, such as sketching buildings in the sky or drawing lamps on walls, as depicted in Fig.~\ref{fig:ade-teaser}(a, b). 
We believe these hallucinations stem from the imcompatibility between the spatial noise prior (which features a mix of class modes after reduction) and ControlNet's control branch (which is only trained to denoise noised tokens of corresponding classes with the label mask). 
This mismatch confuses the control branch, leading to the generation of residuals (added back to the SD branch) that are less effective in denoising the current sample. Such errors accumulate along the denoising trajectory, potentially exacerbating the denoising process further away from the intended trajectory.

Therefore, we delve into the analysis of running statistics on a class-specific basis, calculating a categorical prior for each class. Initially, we downscale $M$ to $M' \in \mathbb{N}^{H' \times W'}$ through nearest pixel selection, employing a scale factor of $\frac{H}{H'}$. 
Subsequently, for $N$ reference images, we organize distinct sets $N_c$ for each class $c$ in the set of classes $C$, which comprise encoded tokens of dimension $1 \times 4$. Following this, we compute the mean and standard deviation for each class to achieve a class-wise statistical reduction.
\begin{equation}
\label{eq:class_prior}
    \mathcal{N}_{\text{categorical}, c} := \mathcal{N}\left( \mathrm{Mean}\left[ N_c \right], \mathrm{Var}\left[N_c\right] \right).
\end{equation}

The definition of $\mathcal{N}_{\text{categorical}, c, \mu T}$ is a replication of Eq.~\eqref{eq:spatial-uT}. 
To examine the knowledge encoded in the categorical prior, we also conduct a case study in Fig.~\ref{fig:categorical-prior}. 
We find that the latent features denoised from the categorical prior achieve a multi-modal understanding of natural languages at the stage of object shaping.
This can explain why our categorical prior can generate results that have better alignment with the given semantic mask.
However, simply using the categorical prior makes the color scheme revert to a monotonous manner.

\subsection{Joint Prior}
In Sec.~\ref{spatial-class-prior}, we explore spatial prior, which aids in constructing scene layouts but falls short in generalizing to class-specific details, and class priors, which excel in generating localized objects of a certain class, yet lack comprehensive global attention. In this section, we introduce a combined prior that effectively merges these two aspects, which we dub as \textbf{joint prior}.

The calculation of the joint prior, $\mathcal{N}_\text{joint}$, can be formulated as follows. For $N$ reference images, we use different sets $N_{x,y,c}$ to store encoded tokens with dimensions $1 \times 4$. 
Subsequently, the mean and variance are calculated for each tuple $(x,y,c)$, where $x\in[0, H'), y\in[0, W'), c\in C$,
\begin{equation}
\label{eq:joint_prior}
    \mathcal{N}_{\text{joint}, x,y,c} := \mathcal{N}\left( \mathrm{Mean}\left[ N_{x,y,c} \right], \mathrm{Var}\left[N_{x,y,c}\right] \right).
\end{equation}

\begin{figure}[!t]
\centering
\includegraphics[width=0.75\textwidth]{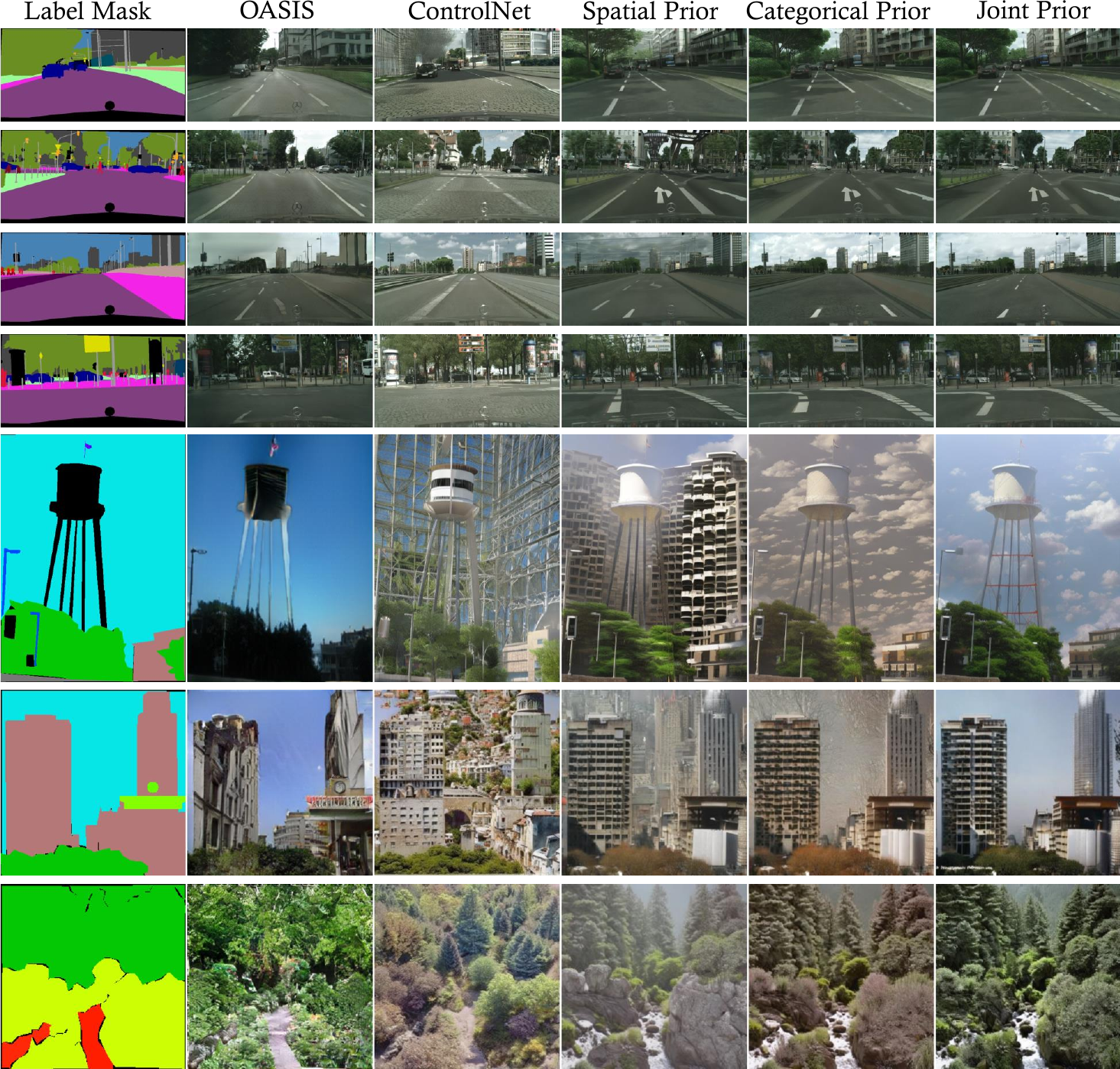}
\caption{\textbf{Qualitative Comparison of Different Noise Priors.} The resolutions for generated results on Cityscapes (top 4 rows) \cite{cordts2016cityscapes} and ADE20K (bottom 3 rows) \cite{zhou2017scene} are $1024 \times 512$ and $512 \times 512$, respectively. Please zoom in for a better view.}
\label{fig:prior-comp}
\end{figure}

In instances where the sample size for a given tuple is excessively small, we revert to taking the statistics from class prior, assuming that such situations are rare and indicating a scenario where the local shape formation of an object should take precedence. The formulation of $\mathcal{N}_{\text{joint}, x,y,c, \mu T}$ is also like Eq.~\eqref{eq:spatial-uT}.
As evidenced by both quantitative (Tab. \ref{table:baseline}) and qualitative analyses (Fig.~\ref{fig:prior-comp}), the joint prior effectively incorporates the strengths of spatial and categorical priors.

\label{joint_prior}
\textbf{Discussion of Denoising steps $\boldsymbol{\mu T}$.}
Ideally, the coefficient $\mu$ of the denoising timesteps needs to be carefully tuned. A smaller $\mu$ means injecting lower levels of noise into the calculated latent priors, thereby reducing the number of denoising steps required and accelerating the inference process.
Nevertheless, within the framework of the joint prior, as we treat encoded tokens from varying spatial positions and categories as independent variables and overlook their correlations, a higher number of denoising steps $\mu$ becomes necessary.
This requirement is due to the need for more self-attention to transition the marginal statistics towards joint modeling progressively.
On a practical note, employing DDIM can expedite this search problem, significantly narrowing down the search space of $\mu$-s to a stride of $0.05$.
We provide this study in Fig.~\ref{fig:uT}.

\begin{table}[!b]
    \centering
    \caption{\textbf{Quantitative Comparison of Different Noise Priors.}}
    \begin{adjustbox}{max width=0.9\linewidth}
    \begin{tblr}{
      cells = {c,},
      cell{1}{1} = {r=2}{},
      cell{1}{2} = {c=3}{c},
      cell{1}{5} = {c=3}{c},
      vline{2, 5} = {}
    }
    \hline[2pt, solid]
    Method & Cityscapes \cite{cordts2016cityscapes} & & & ADE20K \cite{zhou2017scene} & & & \\
           & mIoU $\uparrow$ & Acc $\uparrow$ & FID $\downarrow$ & mIoU $\uparrow$ & Acc $\uparrow$ & FID $\downarrow$ &\\
    \hline
Normal Prior & 65.14$_{\textcolor{black}{(+0.00)}}$ & 94.14$_{\textcolor{black}{(+0.00)}}$ & 23.35$_{\textcolor{black}{(+0.00)}}$ & 20.73$_{\textcolor{black}{(+0.00)}}$ & 61.14$_{\textcolor{black}{(+0.00)}}$ & 20.58$_{\textcolor{black}{(+0.00)}}$ & \\
Spatial Prior  &  66.77$_{(\textcolor{blue}{+1.63})}$ & 94.29$_{(\textcolor{blue}{+0.15})}$ & 12.83$_{(\textcolor{red}{-10.52})}$ &20.86$_{(\textcolor{blue}{+0.13})}$ & 64.46$_{(\textcolor{blue}{+3.32})}$ & 16.03$_{(\textcolor{red}{-4.55})}$ & \\
Categorical Prior & 66.86$_{(\textcolor{blue}{+1.72})}$ & 94.54$_{(\textcolor{blue}{+0.40})}$ & 11.63$_{(\textcolor{red}{-11.72})}$ & 21.86$_{(\textcolor{blue}{+1.13})}$ & 66.63$_{(\textcolor{blue}{+5.49})}$ & 16.56$_{(\textcolor{red}{-4.02})}$ & \\
Joint Prior  & \textbf{67.92}$_{(\textcolor{blue}{+2.78})}$ & \textbf{94.65}$_{(\textcolor{blue}{+0.51})}$ & \textbf{10.53}$_{(\textcolor{red}{-12.82})}$ & \textbf{25.61}$_{(\textcolor{blue}{+4.88})}$ & \textbf{71.79}$_{(\textcolor{blue}{+10.65})}$ & \textbf{12.66}$_{(\textcolor{red}{-7.92})}$ & \\
    \hline[2pt, solid]
\end{tblr}
\end{adjustbox}
    \label{table:baseline}
\end{table}

\section{Experiments}

\subsection{Setup}

We evaluate our proposed noise priors on three challenging datasets: Cityscapes \cite{cordts2016cityscapes}, ADE20K \cite{zhou2017scene} and COCO-Stuff \cite{caesar2018coco}.
We use (i) the mean intersection-over-union (mIoU) and pixel accuracy (Acc) to evaluate the alignment with the provided label mask (following SPADE \cite{park2019semantic}), (ii) the Fréchet inception distance (FID) score \cite{heusel2017gans} to access the quality of generated images, (iii) LPIPS \cite{zhang2018perceptual} and MS-SSIM \cite{wang2003multiscale} to evaluate the diversity of generated images, and (iv) user study to see the aesthetic appeal of the outcomes.

We pre-compute the proposed priors on $N$=10,000 images.
For Cityscapes and ADE20K datasets, these latent priors are applied to ControlNet\cite{zhang2023adding} finetuned on these two datasets, respectively.
This finetuning process is conducted using a batch size of 16 on A100 80G GPUs, employing a learning rate of $10^{-5}$ across 100,000 steps. The decoder in the original Stable Diffusion \cite{rombach2022high} branch is also unlocked for tuning. 
For the COCO-Stuff dataset, we apply our method to the finetuned FreestyleNet model\cite{xue2023freestyle}, which has achieved state-of-the-art performance in the COCO-Stuff semantic image segmentation (SIS) task. We utilize the official checkpoint provided by the authors.

\subsection{Main Results}
\begin{wraptable}[19]{R}{0.55\textwidth}
\caption{
\textbf{Comparison of Our Method with State-of-the-Art Approaches.}}
    \centering
    \resizebox{0.9\linewidth}{!}{
    \begin{tblr}{
      cells = {c,},
      cell{1}{1} = {r=2}{},
      cell{1}{2} = {c=3}{c},
      vline{2} = {}
    }
    \hline[2pt, solid]
    Method & FID & \\
           & Cityscapes & ADE20K & COCO-Stuff & \\
           
    \hline
    CRN \cite{chen2017photographic} & 104.7 & 73.3 & 70.4 & \\
    SIMS \cite{qi2018semi} & 49.7 & -     & -   & \\
    Pix2pixHD \cite{wang2018high}  & 95.0 & 81.8 & 111.5 & \\
    GauGAN \cite{park2019semantic}  & 71.8 & 33.9 & 22.6 & \\
    DPGAN  \cite{tang2021layout} & 53.0 & 31.7 &  -   & \\
    DAGAN \cite{tang2020dual}  & 60.3 & 31.9 &  -   & \\
    SelectionGAN \cite{tang2019multi} & 65.2 & 33.1 & - & \\
    SelectionGAN++ \cite{tang2022multi} & 63.4 & 32.2 & - & \\
    LGGAN \cite{tang2020local}  & 57.7 & 31.6 & - & \\
    LGGAN++ \cite{tang2022local} & 48.1 & 30.5 & - & \\
    CC-FPSE \cite{liu2019learning} & 54.3 & 31.7 & 19.2 & \\
    INADE \cite{tan2021diverse} & 44.3 & 35.2 & -\\
    GroupDNet \cite{zhu2020semantically} & 47.3 & 41.7 & -\\
    SC-GAN \cite{wang2021image} & - & 29.3 & 18.1 \\
    OASIS \cite{sushko2022oasis}  & 47.7 & 28.3 & 17.0 & \\
    RESAIL \cite{shi2022retrieval} & 45.5 & 30.2 & 18.3 & \\
    SAFM \cite{lv2022semantic}   & 49.5 & 32.8 & 24.6 & \\
    ECGAN \cite{tang2020edge} & 44.5 & 25.8 & 15.7 & \\
    \hline
    PITI \cite{wang2022pretraining} & - & 27.9 & 16.1 \\
    FreestyleNet \cite{xue2023freestyle} & - & 25.0 & 14.4 \\
    SDM \cite{wang2022semantic}     & 42.1 & 27.5 & 15.9 & \\
    ControlNet \cite{zhang2023adding} & 23.4 & 20.6 & 28.0 & \\
    \hline
    SCP-Diff (Joint prior) & \textbf{10.5} & \textbf{12.7} & \textbf{11.3} &    \\
    \hline[2pt, solid]
\end{tblr}
}

\label{table:sota}

\end{wraptable}

\textbf{Comparison of Noise Priors.}
In this section, we quantitatively and qualitatively compare different noise priors we proposed in this paper, including normal prior (which reduces to standard finetuned ControlNet \cite{zhang2023adding} inference scheme), spatial prior, categorical prior and joint prior. The results are reported in Tab.~\ref{table:baseline} and Fig.~\ref{fig:prior-comp}.

Tab.~\ref{table:baseline} shows that the joint prior outperforms the standard inference prior of a normal distribution used by ControlNet \cite{zhang2023adding} in terms of image quality (FID) and consistency with the provided label masks (mIoU and Acc).
For instance, on the Cityscapes dataset, our approach registers an improvement in mIoU of \textcolor{blue}{+2.78} and a significant reduction in FID of \textcolor{red}{-12.82}. 
Examining the generated images more closely reveals that, while ControlNet tends to produce images with softer edges and fewer blur effects compared to OASIS, it struggles with scene layout organization and often fails to align correctly with the provided label masks, such as incorrectly placing buildings in the sky area. 
Conversely, our proposed joint prior addresses these issues, significantly enhancing photo-realism.

\textbf{Comparison with State-of-the-Arts}
Compared to state-of-the-art advancements in semantic image synthesis, our proposed joint prior (or SCP-Diff) maintains a competitive edge. 
As shown in Tab.~\ref{table:sota}, ControlNet \cite{zhang2023adding} (or normal prior), by shifting the focus of generative modeling from pixel space to latent space, already significantly surpasses earlier methods, benefiting from the ability to synthesize high-resolution images. 
Our approach amplifies this advantage by addressing the issue of the inherent gap in the inference prior distribution of diffusion models, thus setting an astonishing low FID score for the Cityscapes \cite{cordts2016cityscapes} and ADE20K \cite{zhou2017scene} datasets.
As for the COCO-Stuff dataset, we apply our proposed spatial-categorial joint prior to previous state-of-the-art methods on SIS, FreestyleNet \cite{xue2023freestyle}, and we also witnessed a performance gain of \textcolor{red}{-3.1} in FID.




\subsection{Ablation Studies}

\begin{figure}[h]
    \centering
    \begin{minipage}[b]{0.48\textwidth}
        \centering
        \includegraphics[width=\linewidth]{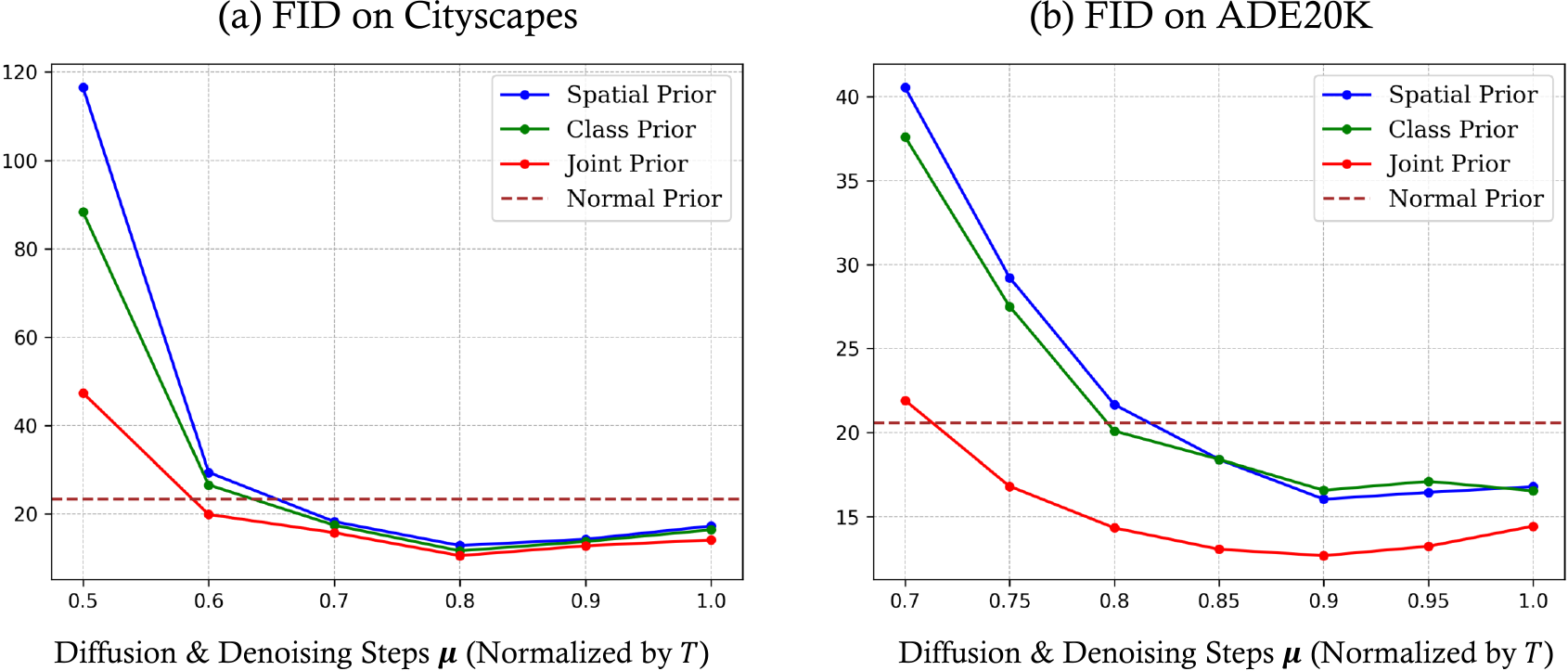}
        \caption{Study on the Effects of Denoising Steps $\mu T$.}
        \label{fig:uT}
    \end{minipage}
    \begin{minipage}[b]{0.48\textwidth}
        \centering
        \includegraphics[width=0.65\linewidth]{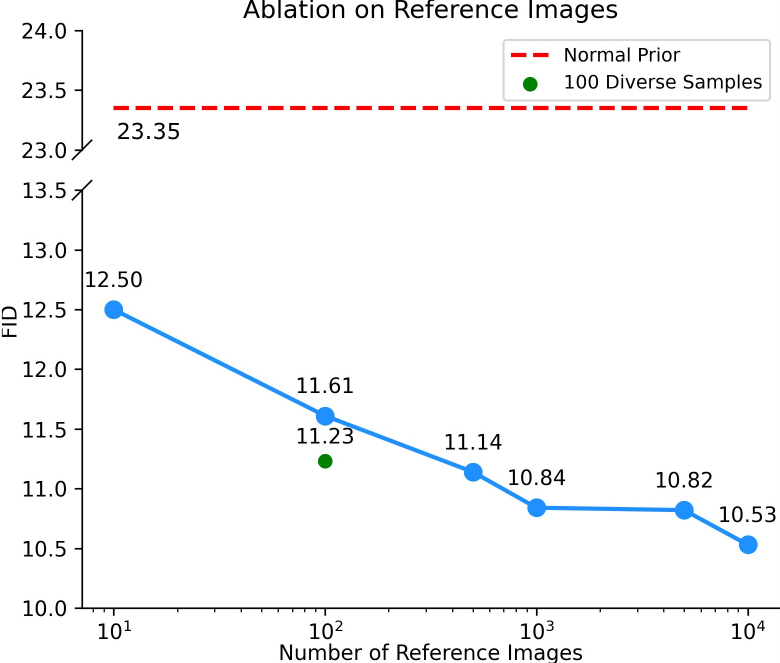}
        \caption{Ablation Study on Number of Reference Images.}
        \label{fig:ablation_on_ref}
    \end{minipage}
\end{figure}

\textbf{Study on Denoising Steps} $\boldsymbol{\mu T}$ By applying inference noise priors at $\mu T$ where $\mu < 1$, we can optimize the traditionally time-consuming sampling method of latent diffusion models. As discussed in Sec.~\ref{joint_prior}, we conducted an experiment to see the effect of $\mu T$ over the quality of the generated results.
From Fig.~\ref{fig:uT}, we can see that the optimal choice of $\mu$ for quality lies in $[0.8, 0.9]$ for both datasets.
The trend patterns observed in both curves are similar, indicating a level of robustness when applying noise priors.





\textbf{Study on Diversity of Generated Results} Following OASIS \cite{sushko2022oasis}, we assess the diversity of the generated images by analyzing the variation within a set of images created from the same label map, referred to as a batch, using MS-SSIM and LPIPS. We produce 20 images per label map, calculate the mean pairwise scores across these images, and then average these scores over the label maps. 
From the results in Tab.~\ref{table:diversity}, we see that our joint prior registers a marginally reduced diversity score compared to ControlNet \cite{zhang2023adding}. This outcome is anticipated, considering that introducing priors to the inference process inherently balances diversity against improved quality.

\begin{table}[!t]
\centering
\begin{minipage}[b]{0.4\linewidth}
\centering
\caption{\textbf{Diversity of Results.}}
\label{table:diversity}
\resizebox{0.85\textwidth}{!}{
\begin{tblr}{
      cells = {c,},
      cell{1}{1} = {r=2}{},
      cell{1}{2} = {c=2}{},
    }
\hline[1.5pt, solid]
Method & ADE20K \cite{zhou2017scene} \\
                        & LPIPS ↑     & MS-SSIM ↓    \\
\hline
OASIS \cite{sushko2022oasis}                   & 0.35        & 0.65         \\
ControlNet \cite{zhang2023adding}              & \textbf{0.59}        & \textbf{0.17}         \\
SCP-Diff                & \underline{0.56}        & \underline{0.21}        \\
\hline[1.5pt, solid]
\end{tblr}
}
\end{minipage}
\hfill
\begin{minipage}[b]{0.55\linewidth}
\caption{\textbf{Results of User Study.}}
\label{table:user_study}
\resizebox{1.00\textwidth}{!}{
\begin{tblr}{
      cells = {c,},
    }
\hline[1.5pt, solid]
Method & Result Quality $\uparrow$ & Condition Fidelity $\uparrow$ & \\
\hline
OASIS \cite{sushko2022oasis}  & 1.45 ± 0.48 & 1.65 ± 0.59 & \\
ControlNet \cite{zhang2023adding} & 1.93 ± 0.46 & 1.80 ± 0.52 \\
SCP-Diff & \textbf{2.62 ± 0.36} & \textbf{2.55 ± 0.43}
\\
\hline[1.5pt, solid]
\end{tblr}
}
\end{minipage}
\end{table}

\textbf{Study on Reference Images} To study the effect of sample size $N$, we present an ablation study in Fig~\ref{fig:ablation_on_ref}. The figure shows that as the sample size increases, the FID score decreases. Even with a limited number of reference images, the FID score is still significantly lower than that of the original ControlNet.
Furthermore, to examine if sample diversity helps, we use the CLIP image encoder (ViT-B/32) to encode 10,000 images and conduct furthest point sampling within the feature space. We select 100 images with the greatest distance and use these images to compute the joint prior. The scatter plot in Fig.~\ref{fig:ablation_on_ref} shows that the diversity of images can be beneficial to the final image quality.



\textbf{User Study} Following ControlNet \cite{zhang2023adding}, we conduct a user study and invite participants to rank 500 groups of images generated by three methods (see Tab.~\ref{table:user_study}) individually in terms of ``the quality of displayed images'' and ``the fidelity to the given label mask''. 
We employ the Average Human Ranking (AHR) as a metric for user preference, where participants rank each result on a scale from 1 to 3, with 3 being the best.
According to the average rankings shown in Tab.~\ref{table:user_study}, users significantly favor the outcomes produced by our method over those generated by ControlNet \cite{zhang2023adding}.

\section{Conclusion}
In this paper, we have addressed the challenge of the inference distribution discrepancy in finetuned ControlNets for SIS by introducing inference noise priors that effectively bridge this gap
Our SCP-Diff showcases outstanding performance, setting new benchmarks in SIS on Cityscapes and ADE20K datasets.
We hope our insights and high-quality generated images will inspire future works in the research community.\footnote{The real images in Fig.~\ref{fig:cs-teaser} are all in the left.}\footnote{This research is supported by Tsinghua University – Mercedes Benz Institute for Sustainable Mobility.}

%


\section*{Acknowledgements}
This research is supported by Tsinghua University – Mercedes Benz Institute for Sustainable Mobility.
%
\bibliographystyle{splncs04}
\bibliography{main}

\begin{thebibliography}{10}
\providecommand{\url}[1]{\texttt{#1}}
\providecommand{\urlprefix}{URL }
\providecommand{\doi}[1]{https://doi.org/#1}

\bibitem{avrahami2022blended}
Avrahami, O., Lischinski, D., Fried, O.: Blended diffusion for text-driven editing of natural images. In: Proceedings of the IEEE/CVF Conference on Computer Vision and Pattern Recognition. pp. 18208--18218 (2022)

\bibitem{balaji2022ediffi}
Balaji, Y., Nah, S., Huang, X., Vahdat, A., Song, J., Kreis, K., Aittala, M., Aila, T., Laine, S., Catanzaro, B., et~al.: ediffi: Text-to-image diffusion models with an ensemble of expert denoisers. arXiv preprint arXiv:2211.01324  (2022)

\bibitem{caesar2018coco}
Caesar, H., Uijlings, J., Ferrari, V.: Coco-stuff: Thing and stuff classes in context. In: Proceedings of the IEEE conference on computer vision and pattern recognition. pp. 1209--1218 (2018)

\bibitem{cao2023masactrl}
Cao, M., Wang, X., Qi, Z., Shan, Y., Qie, X., Zheng, Y.: Masactrl: Tuning-free mutual self-attention control for consistent image synthesis and editing. arXiv preprint arXiv:2304.08465  (2023)

\bibitem{chen2024ultraman}
Chen, M., Chen, J., Ye, X., Gao, H.a., Chen, X., Fan, Z., Zhao, H.: Ultraman: Single image 3d human reconstruction with ultra speed and detail. arXiv preprint arXiv:2403.12028  (2024)

\bibitem{chen2017photographic}
Chen, Q., Koltun, V.: Photographic image synthesis with cascaded refinement networks. In: Proceedings of the IEEE international conference on computer vision. pp. 1511--1520 (2017)

\bibitem{cordts2016cityscapes}
Cordts, M., Omran, M., Ramos, S., Rehfeld, T., Enzweiler, M., Benenson, R., Franke, U., Roth, S., Schiele, B.: The cityscapes dataset for semantic urban scene understanding. In: Proceedings of the IEEE conference on computer vision and pattern recognition. pp. 3213--3223 (2016)

\bibitem{couairon2022diffedit}
Couairon, G., Verbeek, J., Schwenk, H., Cord, M.: Diffedit: Diffusion-based semantic image editing with mask guidance. arXiv preprint arXiv:2210.11427  (2022)

\bibitem{duan2023diffusiondepth}
Duan, Y., Guo, X., Zhu, Z.: Diffusiondepth: Diffusion denoising approach for monocular depth estimation. arXiv preprint arXiv:2303.05021  (2023)

\bibitem{esser2021taming}
Esser, P., Rombach, R., Ommer, B.: Taming transformers for high-resolution image synthesis. In: Proceedings of the IEEE/CVF conference on computer vision and pattern recognition. pp. 12873--12883 (2021)

\bibitem{everaert2024exploiting}
Everaert, M.N., Fitsios, A., Bocchio, M., Arpa, S., S{\"u}sstrunk, S., Achanta, R.: Exploiting the signal-leak bias in diffusion models. In: Proceedings of the IEEE/CVF Winter Conference on Applications of Computer Vision. pp. 4025--4034 (2024)

\bibitem{gao2023semi}
Gao, H.a., Tian, B., Li, P., Chen, X., Zhao, H., Zhou, G., Chen, Y., Zha, H.: From semi-supervised to omni-supervised room layout estimation using point clouds. In: 2023 IEEE International Conference on Robotics and Automation (ICRA). pp. 2803--2810. IEEE (2023)

\bibitem{gao2023dqs3d}
Gao, H.a., Tian, B., Li, P., Zhao, H., Zhou, G.: Dqs3d: Densely-matched quantization-aware semi-supervised 3d detection. In: Proceedings of the IEEE/CVF International Conference on Computer Vision. pp. 21905--21915 (2023)

\bibitem{ge2023preserve}
Ge, S., Nah, S., Liu, G., Poon, T., Tao, A., Catanzaro, B., Jacobs, D., Huang, J.B., Liu, M.Y., Balaji, Y.: Preserve your own correlation: A noise prior for video diffusion models. In: Proceedings of the IEEE/CVF International Conference on Computer Vision. pp. 22930--22941 (2023)

\bibitem{goodfellow2020generative}
Goodfellow, I., Pouget-Abadie, J., Mirza, M., Xu, B., Warde-Farley, D., Ozair, S., Courville, A., Bengio, Y.: Generative adversarial networks. Communications of the ACM  \textbf{63}(11),  139--144 (2020)

\bibitem{heusel2017gans}
Heusel, M., Ramsauer, H., Unterthiner, T., Nessler, B., Hochreiter, S.: Gans trained by a two time-scale update rule converge to a local nash equilibrium. Advances in neural information processing systems  \textbf{30} (2017)

\bibitem{ho2020denoising}
Ho, J., Jain, A., Abbeel, P.: Denoising diffusion probabilistic models. Advances in neural information processing systems  \textbf{33},  6840--6851 (2020)

\bibitem{huang2017arbitrary}
Huang, X., Belongie, S.: Arbitrary style transfer in real-time with adaptive instance normalization. In: Proceedings of the IEEE international conference on computer vision. pp. 1501--1510 (2017)

\bibitem{jiang2024p}
Jiang, Z., Zhu, Z., Li, P., Gao, H.a., Yuan, T., Shi, Y., Zhao, H., Zhao, H.: P-mapnet: Far-seeing map generator enhanced by both sdmap and hdmap priors. arXiv preprint arXiv:2403.10521  (2024)

\bibitem{li2024fairdiff}
Li, W., Xu, H., Zhang, G., Gao, H.a., Gao, M., Wang, M., Zhao, H.: Fairdiff: Fair segmentation with point-image diffusion. arXiv preprint arXiv:2407.06250  (2024)

\bibitem{lin2024common}
Lin, S., Liu, B., Li, J., Yang, X.: Common diffusion noise schedules and sample steps are flawed. In: Proceedings of the IEEE/CVF Winter Conference on Applications of Computer Vision. pp. 5404--5411 (2024)

\bibitem{liu2019learning}
Liu, X., Yin, G., Shao, J., Wang, X., et~al.: Learning to predict layout-to-image conditional convolutions for semantic image synthesis. Advances in Neural Information Processing Systems  \textbf{32} (2019)

\bibitem{lu2019closed}
Lu, M., Zhao, H., Yao, A., Chen, Y., Xu, F., Zhang, L.: A closed-form solution to universal style transfer. In: Proceedings of the IEEE/CVF International Conference on Computer Vision. pp. 5952--5961 (2019)

\bibitem{luo2022context}
Luo, W., Yang, S., Wang, H., Long, B., Zhang, W.: Context-consistent semantic image editing with style-preserved modulation. In: European Conference on Computer Vision. pp. 561--578. Springer (2022)

\bibitem{luo2023siedob}
Luo, W., Yang, S., Zhang, X., Zhang, W.: Siedob: Semantic image editing by disentangling object and background. In: Proceedings of the IEEE/CVF Conference on Computer Vision and Pattern Recognition. pp. 1868--1878 (2023)

\bibitem{lv2022semantic}
Lv, Z., Li, X., Niu, Z., Cao, B., Zuo, W.: Semantic-shape adaptive feature modulation for semantic image synthesis. In: Proceedings of the IEEE/CVF Conference on Computer Vision and Pattern Recognition. pp. 11214--11223 (2022)

\bibitem{lv2024place}
Lv, Z., Wei, Y., Zuo, W., Wong, K.Y.K.: Place: Adaptive layout-semantic fusion for semantic image synthesis (2024)

\bibitem{mokady2023null}
Mokady, R., Hertz, A., Aberman, K., Pritch, Y., Cohen-Or, D.: Null-text inversion for editing real images using guided diffusion models. In: Proceedings of the IEEE/CVF Conference on Computer Vision and Pattern Recognition. pp. 6038--6047 (2023)

\bibitem{ntavelis2020sesame}
Ntavelis, E., Romero, A., Kastanis, I., Van~Gool, L., Timofte, R.: Sesame: Semantic editing of scenes by adding, manipulating or erasing objects. In: Computer Vision--ECCV 2020: 16th European Conference, Glasgow, UK, August 23--28, 2020, Proceedings, Part XXII 16. pp. 394--411. Springer (2020)

\bibitem{park2019semantic}
Park, T., Liu, M.Y., Wang, T.C., Zhu, J.Y.: Semantic image synthesis with spatially-adaptive normalization. In: Proceedings of the IEEE/CVF conference on computer vision and pattern recognition. pp. 2337--2346 (2019)

\bibitem{qi2018semi}
Qi, X., Chen, Q., Jia, J., Koltun, V.: Semi-parametric image synthesis. In: Proceedings of the IEEE Conference on Computer Vision and Pattern Recognition. pp. 8808--8816 (2018)

\bibitem{qiu2023freenoise}
Qiu, H., Xia, M., Zhang, Y., He, Y., Wang, X., Shan, Y., Liu, Z.: Freenoise: Tuning-free longer video diffusion via noise rescheduling. arXiv preprint arXiv:2310.15169  (2023)

\bibitem{ramesh2022hierarchical}
Ramesh, A., Dhariwal, P., Nichol, A., Chu, C., Chen, M.: Hierarchical text-conditional image generation with clip latents. arXiv preprint arXiv:2204.06125  \textbf{1}(2), ~3 (2022)

\bibitem{rombach2022high}
Rombach, R., Blattmann, A., Lorenz, D., Esser, P., Ommer, B.: High-resolution image synthesis with latent diffusion models. In: Proceedings of the IEEE/CVF conference on computer vision and pattern recognition. pp. 10684--10695 (2022)

\bibitem{shi2022retrieval}
Shi, Y., Liu, X., Wei, Y., Wu, Z., Zuo, W.: Retrieval-based spatially adaptive normalization for semantic image synthesis. In: Proceedings of the IEEE/CVF Conference on Computer Vision and Pattern Recognition. pp. 11224--11233 (2022)

\bibitem{song2020denoising}
Song, J., Meng, C., Ermon, S.: Denoising diffusion implicit models. arXiv preprint arXiv:2010.02502  (2020)

\bibitem{sushko2022oasis}
Sushko, V., Sch{\"o}nfeld, E., Zhang, D., Gall, J., Schiele, B., Khoreva, A.: Oasis: only adversarial supervision for semantic image synthesis. International Journal of Computer Vision  \textbf{130}(12),  2903--2923 (2022)

\bibitem{tan2021diverse}
Tan, Z., Chai, M., Chen, D., Liao, J., Chu, Q., Liu, B., Hua, G., Yu, N.: Diverse semantic image synthesis via probability distribution modeling. In: Proceedings of the IEEE/CVF Conference on Computer Vision and Pattern Recognition. pp. 7962--7971 (2021)

\bibitem{tan2021efficient}
Tan, Z., Chen, D., Chu, Q., Chai, M., Liao, J., He, M., Yuan, L., Hua, G., Yu, N.: Efficient semantic image synthesis via class-adaptive normalization. IEEE Transactions on Pattern Analysis and Machine Intelligence  \textbf{44}(9),  4852--4866 (2021)

\bibitem{tang2020dual}
Tang, H., Bai, S., Sebe, N.: Dual attention gans for semantic image synthesis. In: Proceedings of the 28th ACM International Conference on Multimedia. pp. 1994--2002 (2020)

\bibitem{tang2020edge}
Tang, H., Qi, X., Sun, G., Xu, D., Sebe, N., Timofte, R., Van~Gool, L.: Edge guided gans with contrastive learning for semantic image synthesis. arXiv preprint arXiv:2003.13898  (2020)

\bibitem{tang2021layout}
Tang, H., Sebe, N.: Layout-to-image translation with double pooling generative adversarial networks. IEEE Transactions on Image Processing  \textbf{30},  7903--7913 (2021)

\bibitem{tang2022local}
Tang, H., Shao, L., Torr, P.H., Sebe, N.: Local and global gans with semantic-aware upsampling for image generation. IEEE Transactions on Pattern Analysis and Machine Intelligence  \textbf{45}(1),  768--784 (2022)

\bibitem{tang2023edge}
Tang, H., Sun, G., Sebe, N., Van~Gool, L.: Edge guided gans with multi-scale contrastive learning for semantic image synthesis. IEEE Transactions on Pattern Analysis and Machine Intelligence  (2023)

\bibitem{tang2022multi}
Tang, H., Torr, P.H., Sebe, N.: Multi-channel attention selection gans for guided image-to-image translation. IEEE Transactions on Pattern Analysis and Machine Intelligence  \textbf{45}(5),  6055--6071 (2022)

\bibitem{tang2019multi}
Tang, H., Xu, D., Sebe, N., Wang, Y., Corso, J.J., Yan, Y.: Multi-channel attention selection gan with cascaded semantic guidance for cross-view image translation. In: Proceedings of the IEEE/CVF conference on computer vision and pattern recognition. pp. 2417--2426 (2019)

\bibitem{tang2020local}
Tang, H., Xu, D., Yan, Y., Torr, P.H., Sebe, N.: Local class-specific and global image-level generative adversarial networks for semantic-guided scene generation. In: Proceedings of the IEEE/CVF conference on computer vision and pattern recognition. pp. 7870--7879 (2020)

\bibitem{tian2023unsupervised}
Tian, B., Liu, M., Gao, H.a., Li, P., Zhao, H., Zhou, G.: Unsupervised road anomaly detection with language anchors. In: 2023 IEEE international conference on robotics and automation (ICRA). pp. 7778--7785. IEEE (2023)

\bibitem{voynov2023p+}
Voynov, A., Chu, Q., Cohen-Or, D., Aberman, K.: $ p+ $: Extended textual conditioning in text-to-image generation. arXiv preprint arXiv:2303.09522  (2023)

\bibitem{wang2022pretraining}
Wang, T., Zhang, T., Zhang, B., Ouyang, H., Chen, D., Chen, Q., Wen, F.: Pretraining is all you need for image-to-image translation. arXiv preprint arXiv:2205.12952  (2022)

\bibitem{wang2018high}
Wang, T.C., Liu, M.Y., Zhu, J.Y., Tao, A., Kautz, J., Catanzaro, B.: High-resolution image synthesis and semantic manipulation with conditional gans. In: Proceedings of the IEEE conference on computer vision and pattern recognition. pp. 8798--8807 (2018)

\bibitem{wang2022semantic}
Wang, W., Bao, J., Zhou, W., Chen, D., Chen, D., Yuan, L., Li, H.: Semantic image synthesis via diffusion models. arXiv preprint arXiv:2207.00050  (2022)

\bibitem{wang2021image}
Wang, Y., Qi, L., Chen, Y.C., Zhang, X., Jia, J.: Image synthesis via semantic composition. In: Proceedings of the IEEE/CVF International Conference on Computer Vision. pp. 13749--13758 (2021)

\bibitem{wang2003multiscale}
Wang, Z., Simoncelli, E.P., Bovik, A.C.: Multiscale structural similarity for image quality assessment. In: The Thrity-Seventh Asilomar Conference on Signals, Systems \& Computers, 2003. vol.~2, pp. 1398--1402. Ieee (2003)

\bibitem{wu2023freeinit}
Wu, T., Si, C., Jiang, Y., Huang, Z., Liu, Z.: Freeinit: Bridging initialization gap in video diffusion models. arXiv preprint arXiv:2312.07537  (2023)

\bibitem{xue2023freestyle}
Xue, H., Huang, Z., Sun, Q., Song, L., Zhang, W.: Freestyle layout-to-image synthesis. In: Proceedings of the IEEE/CVF Conference on Computer Vision and Pattern Recognition. pp. 14256--14266 (2023)

\bibitem{yang2024freemask}
Yang, L., Xu, X., Kang, B., Shi, Y., Zhao, H.: Freemask: Synthetic images with dense annotations make stronger segmentation models. Advances in Neural Information Processing Systems  \textbf{36} (2024)

\bibitem{zhang2024preserving}
Zhang, J., Chang, S.Y., Li, K., Forsyth, D.: Preserving image properties through initializations in diffusion models. In: Proceedings of the IEEE/CVF Winter Conference on Applications of Computer Vision. pp. 5242--5250 (2024)

\bibitem{zhang2023adding}
Zhang, L., Rao, A., Agrawala, M.: Adding conditional control to text-to-image diffusion models. In: Proceedings of the IEEE/CVF International Conference on Computer Vision. pp. 3836--3847 (2023)

\bibitem{zhang2018perceptual}
Zhang, R., Isola, P., Efros, A.A., Shechtman, E., Wang, O.: The unreasonable effectiveness of deep features as a perceptual metric. In: CVPR (2018)

\bibitem{zheng2023steps}
Zheng, Y., Zhong, C., Li, P., Gao, H.a., Zheng, Y., Jin, B., Wang, L., Zhao, H., Zhou, G., Zhang, Q., et~al.: Steps: Joint self-supervised nighttime image enhancement and depth estimation. arXiv preprint arXiv:2302.01334  (2023)

\bibitem{zhou2017scene}
Zhou, B., Zhao, H., Puig, X., Fidler, S., Barriuso, A., Torralba, A.: Scene parsing through ade20k dataset. In: Proceedings of the IEEE conference on computer vision and pattern recognition. pp. 633--641 (2017)

\bibitem{zhu2020semantically}
Zhu, Z., Xu, Z., You, A., Bai, X.: Semantically multi-modal image synthesis. In: Proceedings of the IEEE/CVF conference on computer vision and pattern recognition. pp. 5467--5476 (2020)

\end{thebibliography}

\end{document}